\documentclass[11pt,a4paper]{article}

\usepackage[utf8]{inputenc}
\usepackage{algorithm}
\usepackage{algorithmic}
\usepackage{tabularx}
\usepackage{amsmath}
\usepackage{authblk}
\usepackage{graphicx}

\newtheorem{definition}{Definition}
\newtheorem{assumption}{Assumption}
\algsetup{indent=2em}

\usepackage[nodisplayskipstretch]{setspace}
\begin{document}

\title{\emph{x.ent}: R Package for Entities and Relations Extraction based on Unsupervised Learning and Document Structure}

\author[1]{Nicolas Turenne\thanks{nturenne.inra@yahoo.fr}}
\author[2]{Tien Phan\thanks{ptgtien@vnua.edu.vn}}

\affil[1]{INRA \& Universit\'e Paris-Est\\ Laboratoire Interdisciplinaire Sciences Innovations Soci\'et\'es (LISIS), UMR 1326\\ 77430 Champs-sur-Marne, France}
\affil[2]{Vietnam National University of Agriculture\\ Faculty of Information Technology\\ Gialam district, Hanoi, Vietnam}
\providecommand{\keywords}[1]{\textbf{\textit{Keywords:}} #1}

\maketitle
 
\begin{abstract}
Relation extraction with accurate precision is still a challenge when processing full text databases. We propose an approach based on cooccurrence analysis in each document for which we used document organization to improve accuracy of relation extraction. This approach is implemented in a R package called \emph{x.ent}. Another facet of extraction relies on use of extracted relation into a querying system for expert end-users. Two datasets had been used. One of them gets interest from specialists of epidemiology in plant health. For this dataset usage is dedicated to plant-disease exploration through agricultural information news. An open-data platform exploits exports from \emph{x.ent} and is publicly available.
\end{abstract}



\section[Introduction]{ Introduction }

More than 90\% of available information during history have been produced only the last 5 years (see \cite{httpScienceDaily}). All kind of data is concerned but if we consider data on internet, usage considered by people are mostly associated to textual data (i.e sending a tweet or an email)(see \cite{httpWeb}), and on internet 3\% out of 48 billions URLs are indexed by Google; it means 14 billions webpages are under textual format (see \cite{httpFactshunt}). Even on a database such as Youtube we can find text annotations for videos. Text processing gets more and more interest in database processing to sift interesting pieces of information (see \cite{Konchady:2008} \cite{Turenne:2015}). Natural Language Processing is naturally engaged in analytics and for lots of purposes, relations makes sense in many documents (see \cite{Soderland:1999}); not only bags of words as it has been widely studied and implemented for a long time (see \cite{Riloff:1996} \cite{Hearst:1992}). \\
Words are ambiguous syntactically or semantically, but if we consider such information extraction task as relation extraction with named entities it could increase accuracy of extraction because named entities are less ambiguous than noun phrases (see \cite{Turenne:2013}). In lots of specialized domains, users play an important role. The domain acts as a constraint to the lexical universe, and end-users act as a validation process for extracted information. That is why involving users (i.e. domain-experts) in specification of usefulness about extraction add constraints about extracted information by a system. \\
In this sense our proposed system takes into account hand-crafted definition of external resources and user interface specification by domain experts. These resources are of two kinds: the first one is a proto-ontology of domain named entities, the second is rules definition with lexical markers in the case concepts can not fit to usual case of named entities (such as an opinion, a stage ...). Sometimes we can map to a concept both approaches (dictionaries and rules). Once instances of concept are retrieved, we used heuristics of document architecture as rules to reinforce an unsupervised approach by co-occurrence analysis. Our tool x.ent integrates some graphical facilities help a user to explore exported lists of extractions.\\
Part 2 (state of the art) presents the context of named entity recognition, relation extraction of named entities. Part 3 (methodology) explains our approach to extract relations in a document (datasets, heuristics and algorithms). Finally the part 4 (results) shows result assessment and some means to explore relations.

\section[State of the art]{ State of the art }

The field of "Information Extraction" has been defined to identify pieces of information associated to real-world objects such that they can be classified by type, for instance  {organization:org, person:per, localisation:loc, protein:prot, disease:dis, prices:pri, times:tim...} ; and extended recently with others popular components as species, phenotypes, medicines.  This an example :\\ "We performed exome sequencing in a family with \textless Crohn's disease:dis\textgreater (CD) and severe autoimmunity, analysed immune cell phenotype and function in affected and non-affected individuals, and performed in silico and in vitro analyses of  \textless cytotoxic T lymphocyte-associated protein 4:prot\textgreater (CTLA-4) structure and function." \\
Entity recognition is the first stage necessary for other stages that could association of other numeric or qualitative information about context of a named entity, relation extraction between entities, association of group of relations occurring in a same time or a same location, organizing a scenario of item-sets over time or causality. This is some example of relation: involved\_in, located\_in, part\_of, married\_to, sold\_to, interact\_with, causes\_damage\_to...). From the previous example we can set the relation between categories \textless prot\textgreater and \textless dis\textgreater as "\textless prot\textgreater involved\_in\textless dis\textgreater" and the identified instances \textless cytotoxic T lymphocyte-associated protein 4:prot\textgreater \textless Crohn's disease:dis\textgreater fits well. At an upper level of extraction we should able to achieve a network reconstruction or figuring out sets of events. But at this stage, mapping a sequence of events that have not been validated experimentally, or by an expertise, should be only putative. \\
There are two families of approaches. Handcraft-design rules by expert approach is considered as more qualitative or symbolic (also called knowledge-rich), and the machine learning approach is considered as more numeric or quantitative (also called knowledge-poor). But in any case experts are involved in the loop to define resources (well defined lexical dictionaries, annotated files, detection rules). Some learning techniques can avoid intervention of experts as distant learning and raw cooccurrence analysis, but they are not tractable in any case with good accuracy.

Named entity recognition (NER) is considered as a solved problem. First emblematic conferences, on the topic, was held between 1987 and 1997 to identify terrorist activities in news (Message understanding Conferences , see \cite{httpMUC:2014} ). Lasts conferences was about molecular biology information organized by the BioCreativ consortium in 2008, and about information from news at the computational natural language learning confernce (CONLL) in 2003. Look at Section~\ref{subsec:eval} to see the definition of parameters (P or precision, R or recall, F-score) to assess quality of extraction. The two main approaches to recognize named entities are: 

\begin{itemize}

\item Pattern-based algorithms where experts define handcrafted rules. \\
Rule-based techniques often works with automata theory (at least when processing textual data). An automaton is a machine in which a set of states Q contains only a limited number of components and is called a finite-state machine (FSM). FSMs are a set of abstract machines consisting in a set of states (set S), a set of input events (ensemble $\Sigma$), a set of output events (set Z) and a transition state function. The transition state function takes the current state and a event as input and return a new set of events as output and the next state. Hence, it can be seen as a function that maps an ordered sequence of events as input to a corresponding sequence, or a set of events, as output. With such transition function $\Sigma \to Z$, this is the mathematical model which gives formal definitions, a finite-state machine is a quintuple where  : 

\begin{itemize}

\item $\Sigma$ is an alphabet as input (a finite set, non null of symbols).
\item S is a non-null set of states. 
\item $s_0$ is an initial state, an element of S.
\item $\delta$ is a state transition function : $\delta:S\times\Sigma  $ (in the case of a finite state automaton non-determinist it should be  $\delta:S\times\Sigma $ , i.e., $\delta$  should return a set of state).
\item F is a set of final states, a sub-set (maybe null) of S.

\end{itemize}

For determinist and non-determinist FSMs, it is conventional to consider $\delta$ as a partial function, i.e. $\delta(q,x)$ does not have to be defined for all combinations of $q \in S$ et $x\in \Sigma$. If a FSM M is in a state q, the next symbol is x and $\delta(q,x)$ is not defined, then M can export an error (i.e. reject an input). A finite state machine is a limited Turing machine where the reader header produces operations, and still go ahead, from left to right. A FSM can always be represented by a graphical model in which states and transitions are displayed. For instance the following, linear, expression can recognize some names of universities : \\
\hspace *{1cm} [A-Z][a-z]+([ ][A-Z][a-z]+)+University \\
Result could be "Paris Est University" or "New York City University". \\
Dictionary-based approach like in Lingpipe (see \cite{Carpenter:2007}) or LAITOR (Literature Assistant for Identification of Terms co-Occurrences and Relationships) (see \cite{Barbosa-Silva:2010})  can achieve a better score if entities are already stored in external resources. Some systems can used rules with predefined patterns using lexical tokens and grammatical conditions. Basic system using rules are based on finite automata and regular languages (see \cite{McCulloch:1943}, \cite{Mealy:1955}, \cite{Moore:1956},\cite{Hopcroft:1979}). More complex system use frames (see \cite{Black:1998}).

\item Statistical Learning algorithms: HMM (see \cite{httpBaumWelch} \cite{httpViterbi} \cite{Appelt:1993} \cite{Bikel:2005}), MAXENT (see \cite{Berger:1996} \cite{Borthwick:1999}, \cite{McCallum:2000}) , CRF (see \cite{Lafferty:2001}), regularized averaged perceptron. These approaches require annotated files to learn dependencies probabilities for the model. IllinoisNER system achieves 90.6\% F1 and SNER (Stanford NER) 86.86\% F1 to the CoNLL03 NER shared task data. IllinoisNER use a perceptron model and SNER a MaxEnt approach. HIT LTP NER use also a MaxEnt and obtain 92.25\% F1 at  OpenNLP namefinder (see \cite{Hornik:2014})  uses also MAxEnt approach. 
\end{itemize}

General accuracy evaluation challenges give a F-score \textgreater  90\% for NER task when at the time human assessment is $\approx$ 97\%.
Relation Extraction  is more difficult (generally F-score \textless  60\%). This is the main basic techniques for relation extraction subdivided into two main approaches as for entities extraction:

\begin{itemize}

\item Exact Dictionary-Based Chunking (see \cite{Chun:2005})
\item Handcrafted rule-based techniques (\cite{Krupka:1995})

\end{itemize}

And optimization approaches :

\begin{itemize}

\item Inductive-Logic Programming (see \cite{Huffman:1996}, \cite{Califf:1999})
\item Bayesian Network Analysis (see \cite{Roth:2002}, \cite{Peshkin:2003}, \cite{Kumar:2014}) 
\item unsupervised learning as co-occurrence analysis (see \cite{Ozgur:1996}, \cite{Blaschke:1999},\cite{Jenssen:2001}, \cite{Nielsen:2005}, \cite{Wong:2009}, \cite{Feinerer:2014})
\item semi-supervised learning, as distant learning (see \cite{Riloff:1993}, \cite{Brin:1998}, \cite{Craven:1999}, \cite{Agichtein:2000}, \cite{Mintz:2009}, \cite{Krieger:2014})

\end{itemize}

In the range of tools using predefined schemas, the tool Pharmspresso (see \cite{Garten:2009}) aims at extracting instances matching with the schemas  "{medicine} {association} {gene}" in full texts documents. Out of 178 genes mentions, Pharmspresso finds 78.1\% and out of 191 medicine mentions the tool finds 74.4\%. With the previously cited schema the tool finds  50.3\% of associations ;  PPInterFinder (see \cite{Raja:2013}) is a tool implemented to extract causal relationships between human proteins in texts relying on 11 schemas. The tool achieves a score 66.05\% on the AIMED corpus and overcomes most of other sytems. Biocreative challenge II  (see \cite{Krallinger:2008}) was focused on molecule name detection and their relationships ; the overview highlights that one of the best scores (F-score=45\%) is given by a system implementing lexical schemas about relationships. OpenDMAP is one of these tools (see \cite{Baumgartner:2008}). \\
From the side of tools implementing symbolic learning, LIEP system outputs 85.2\% F-score (recall 81.6\%; precision 89.4\%) with a training dataset about 100 sentences from the Wall Street Journal. Distant learning gives a precision about 67\% (in a lexical framework) which does not varies with syntactic assumptions (68\%) for 1000 relationships instances concerning films, geography, localisation, persons. A hybrid approach like Sprout-Dare gives 83\% as precision on medical full texts in German. \\
Among statistical learning approaches, the MaxEnt method has been largely used to extract relationships of proteins with a 45\% F-score.
Unsupervised methods based on a simple cooccurrence analysis obtain not so much ridiculous scores (but on abstracts). For instance studies about instance of a schema such as "gene-associated to-disease" highlight a relevance of detection in abstracts of 78.5\% (Precision) and 87.1\% (Recall) (with gene recognition scores P=89\% and R=90.9\%, and about diseases recognition scores P = 90\% and R=96.6\%).

In the era of internet, lots of resources are now available on internet. Generic databases for common knowledge collect millions of facts such as Freebase (see \cite{httpFreebase:2014}) where we can find a relation song/singer like Yesterday  <->  John Lennon, Paul McCartney. Freebase claims to register 3 billions entities with links. Another famous database is Yago (see \cite{httpYago:2014}) storing 10 millions entities and 120 millions of facts. Others generic database can be mentioned and are also parsed by these cited one but individually they can contain lists of interesting pieces of information (WordNet, Wikipedia (see \cite{httpPest:2014}, \cite{httpWheat:2014})). Specific databases propose a summary of a domain like Gene Ontology (GO) in molecular biology or Agrovoc (created the by Food and Agriculture Organization of the United Nations - FAO) in agricultural sciences. These databases are not sufficiently exhaustive even if they can provide lots of relevant relations if we look at conferences such as BioCreative (see \cite{httpBiocreative:2014}).

\section[Methodology]{ Methodology }

\subsection[Datasets]{ Assumptions }
We settle two assumptions and we will assess these assumptions with precision and recall parameters. Our main assumption points out the issue to detect relations when features do not occur only inside a sentence but between sentences and also when a document have some topical organization with titles, subtitles and paragraphs.

\begin{assumption}  \textup{ Architecture of a document  }  \\ 
\label{a1}
\textup{ Relevant items and their relationships can occur to different architectural part of a document  }
\end{assumption} 

Our second assumption is that relationship between named entities are explicit and able to be catched by shallow parsing of tokens in the text.

\begin{assumption} \textup{ Shallow parsing  }  \\ 
\label{a2}
\textup{ Named entities can be captured by shallow parsing }
\end{assumption} 

And a last assumption is that a user can assess results :

\begin{assumption} \textup{ User needs  }  \\ 
\label{a3}
\textup{ Items found in texts can be interpreted by an end-user and fits its usage needs }
\end{assumption} 

With these assumptions we try to infer all possible relationships mentioned by a document. Hence after document segmentation three steps can be interesting to insert into a pipe : \\
\begin{enumerate}
   \item reformatting and segmentation of a document into subparts;
   \item named entity recognition and contextual information extraction.
   \item relation extraction
   \item contextual information assignment (for instance: pest significance, development stage, climat, location).
\end{enumerate}

To describe the problem in a formal way, we introduce the following definitions:

\begin{definition} \textup{ Text Unit and Entity  }  \\ 
\label{d1}
\textup{ A text unit U is a linked list which consists of words W and entities E. An entity can be a word or a set of consecutive words. Entities in a text unit are labeled as E1, E2 ... according to their order, and they take value that range over a set of entity types $C^E$. }
\end{definition} 

Segmentation into text units is not always an easy task because of text components occurring everywhere on a page and not linked sequentially with other text units around. Sometimes a conversion of pdf into ascii format may lead to a merge a text unit into another one. As example with our dataset, on the Figure~\ref{figure_bsv}, we can see that a text unit should start from "Bl\'e" and should end  with "mais ils ne peuvent enrayer les fortes infestations". In this Text unit E1="Bl\'e", E2="bl\'e", E3="orge de printemps", E4="pi\'etin \`echaudage", E5="c\'er\'eale", E6="champignon", E7="maladie", E8="ma\"{i}s", E9="ray-grass", E10="luzerne", E11="soja", E12="champignon", E13="maladie". They are two kinds of named entity $C1=\{E1,E2,E3,E5,E8,E9,E10\}$ $C2=\{E4,E6,E7,E12,E13\}$. C1 represents the "crop" category, C2 represents the "disease" category.

\begin{definition} \textup{ Relation  }  \\ 
\label{d2}
\textup{ A (binary) relation $R_{ij}=(E_i;E_j)$ represents the relation between two entities where $E_i$ is the first argument and $E_j$ is the second. Such relation can belong to a relation type over $C^R$. }
\end{definition} 
As example with our dataset, on the Figure~\ref{figure_bsv}, we can see that the relevant relations are associated with E1="Bl\'e" with entities about concept C2, some $R=\{R_{1.4}, R_{1.6}, R_{1.7}, R_{1.12}, R_{1.13}\}$.
\begin{definition} \textup{ Types of relations between entities  }  \\ 
\label{d3}
\textup{Let denote the predefined set of relations and of entities $C^R$ and $C^E$ respectively. }
\end{definition} 
Lots of kinds of entities can found in  $C^E$, named entities like : pests or crops, but not only, for instance "developmental stage about crops", "developmental stage for pests", or "kinds of damage". About relations, several kinds of relations can be found, for instance \\$C^R=\{ \textit{damage relation between a crop and a disease},\\ \hspace *{1cm} \textit{damage relation between a crop and a pest}, ...\}$.

\subsection[Datasets]{ Datasets }
We call ROMEO the first dataset. It consists of the 5 acts (files) of the classical piece of theater from W. Shakespeare Romeo and Juliet in English. For this dataset the goal of a user needs could be to follow the relationships of persons along the scenes.
We summarize the type of concepts and relations by these ensembles:\\
\hspace *{1cm} $C^E=\{ character \}$ \\
\hspace *{1cm} $C^R=\{ \textit{character talking about or with another character} \}$

We know that \#$C^E=22$ for ROMEO dataset. Number of words is 26,551. Number of tokens (unique words) is 5,846.\\
We call BSV the second dataset. This dataset is a collection of scanned and digital newsletter written in French. The neswletter is published since 1946 but majority of numbers are shared by the French National Library (Biblioth\`eque Mitterrand, or BNF) only since 1963. It is written in each French region weekly to inform about damage on local crops. We work with a sample of 2,323 files. But the dataset in construction should contain about 60,000 files in the range 1963-2015. Each file contain between 1 and 10 pages, 3 in average. For this dataset 8 concepts have been identified manually with experts for which we can design dictionaries. Other kinds information about context for a crop and its relationships should also be of interest such as "developmental stage of a crop", "number of a newsletter", "degree of damage", "climate". We summarize types of concepts and relations by the following ensembles: \\
\hspace *{1cm} $C^E=\{ crops, diseases, pests, auxiliaries, region, towns , chemicals, date, \\ 
\hspace *{2cm}  \textit{developmental stage of a pest}, \textit{developmental stage of a crop},\\  
\hspace *{2cm} \textit{number of a newsletter}, \textit{degree of damage}, climate  \}$ \\
\hspace *{1cm} $C^R=\{ \textit{damage relation between a crop and a disease}, \\ 
\hspace *{2cm} \textit{damage relation between a crop and a pest}, \\ 
\hspace *{2cm} \textit{intensity of damage with relation between a crop and a disease}, \\ 
\hspace *{2cm}  \textit{intensity of damage with relation between a crop and a pest} \}$

\begin{figure}[H]
\centering { \includegraphics[width=350pt]{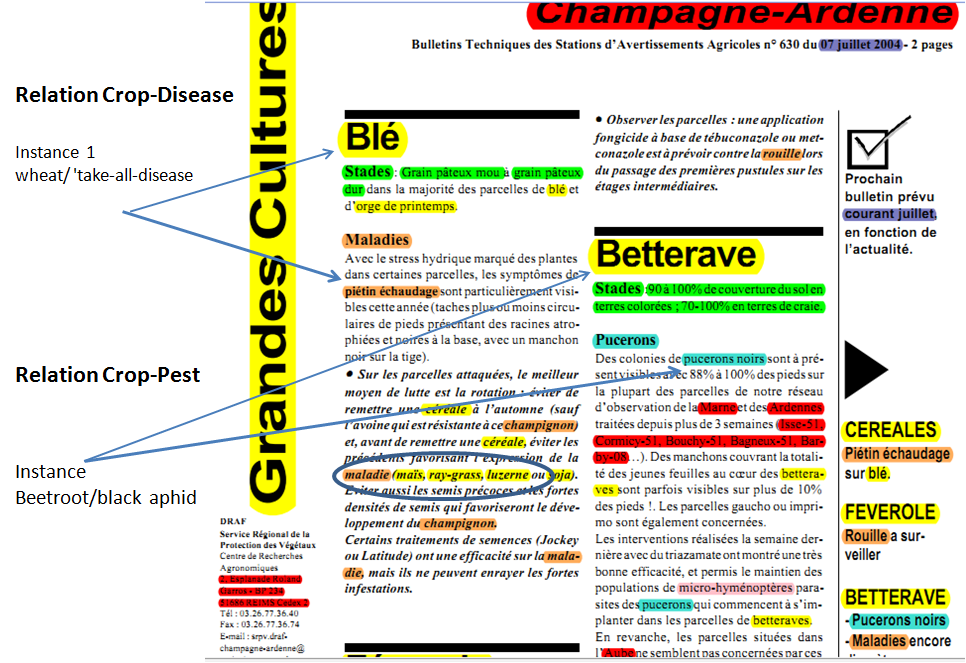} }
\caption{Manual annotations in a document from BSV dataset (in yellow: crops, in green: developmental stages of crops, in brown: diseases, in red: location, in blue: pests, in purple: auxiliaries, in dark blue: time).}
\label{figure_bsv}
\end{figure}

\begin{figure}[H]
\centering { \includegraphics[width=350pt]{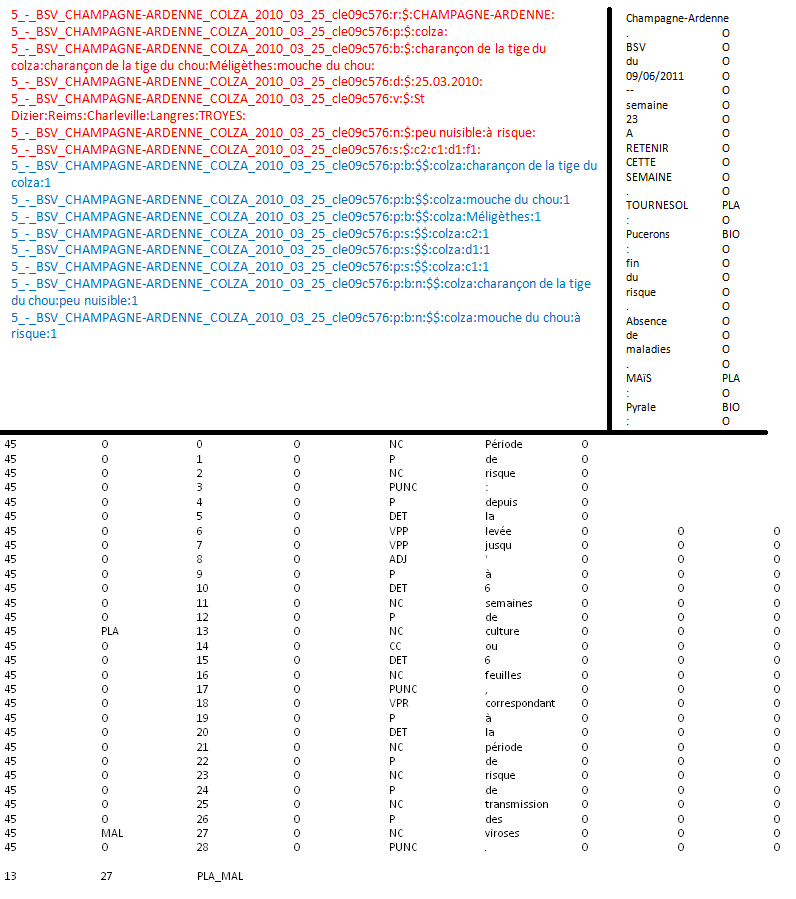} }
\caption{Manual annotations of a document from BSV dataset for evaluation (up-left: csv-like format (in red: entities, in blue: relationships), up-right: BIO/BILOU format for named entities, down: BIO/BILOU format for named entity relationships).}
\label{figure_doc_annot}
\end{figure}

An auxiliary is an insect (as a pest) but not agressive to the crop where it lives. Sometimes it can help control of pests. On the Figure~\ref{figure_bsv} we can see that we can retrieve useful information. Among them relationships crop-disease and crop-disease are also of interest. Thesea are relations pointing agression. As we see in document these relationships does not use verb of other linguistic patterns. \\
To enable evaluation computing we made a annotated dataset about 37 files. To have in mind how much should cost the annotation process, 1000 documents would require 5 months for one person. And as one expert is not sufficient, another control by an extern expert would need more time.
On the Figure~\ref{figure_doc_annot} we see at top-left a sample of manual annotation for a file with nouns phrases denotating concepts (in blue) and relations between these concept (in red). Concepts are cited as a list but a relation occurs only inline. At top-right we see an example of annotation used for CONLL conference challenges about named entity recognition, at bottom the annotation format of CONLL for relation extraction evaluation.

\subsection[Dico]{ Dictionary Matching }

From several concepts denotating named entities about location, persons or biological entities like characters, crops, diseases or regions, we are able to define lexical nouns phrases and a list of entries associated to a set of noun sphrases. These nouns phrases can occur anywhere in the dataset.The format we have adopted can described a hierarchy of entries and lexical variants of each entry. The computing format is a csv-like format. Each line describe an unique entry, followed by an label N (node)or L (leaf) if the entry describe a category or a simple concept. For instance in crops wheat is a species and will be defined as a node; but durum wheat, buckwheat and soft wheat are defined as leaf because they are varieties and linked to wheat. After the category N or L all following nouns phrases are considered as equivalent and could be found in a document. Hence the dictionary collect all relevant entries of a concept describing hypernym relation and synonym relation between lexical phrases. \\
For instance, on the Figure~\ref{figure_dico} we can see a sample of the crop dictionary in French ("bl\'e" is wheat, "bl\'e dur" is buckwheat, "bl\'e tendre" is soft wheat). Figure~\ref{figure_statdico} describe the lexical population for each dictionary about the BSV dataset. About ROMEO dataset we can find 22 characters.
\begin{figure}[H]
\centering
\renewcommand{\arraystretch}{1.2}
{\scriptsize

\begin{tabularx}{12.5cm}{p{12.1cm}}

   \hline
\textbf{bl\'e}:N:bl\'e:BLE:bl\'es:Triticum:bl\'e dur:bl\'e tendre: \\
   \hline
\textbf{bl\'e dur}:L:BLE DUR:T. durum:Triticum durum:bles durs:bl\'es durs:bl\'e dur: \\
   \hline
\textbf{bl\'e noir}:L:BLE NOIR:f. esculentum:fagopyrum esculentum:sarrasin:bles noirs:bl\'es noirs:bl\'e noir:sarrasins: \\
   \hline
\textbf{bl\'e tendre}:L:BLE TENDRE:T. aestivum:Triticum aestivum:bl\'e froment:bl\'es froments:ble froments:bl\'e tendre:bl\'es tendres:bles tendres: \\
   \hline
\end{tabularx}   } 
\caption{Sample of Crops dictionary.}
\label{figure_dico}
\end{figure}

\begin{figure}[H]
\centering
\renewcommand{\arraystretch}{1.2}
{\scriptsize
   \rule{\textwidth}{1pt}
    \textbf{Entities Types}   \\

\begin{tabularx}{12.5cm}{|>{\centering\hsize=0.7\hsize\arraybackslash}X|>{\centering\hsize=0.8\hsize\arraybackslash}X|>{\centering\arraybackslash}X|>{\centering\hsize=0.8\hsize\arraybackslash}X|>{\centering\hsize=1.\hsize\arraybackslash}X|>{\centering\arraybackslash}X|>{\centering\hsize=0.5\hsize\arraybackslash}X|>{\centering\hsize=1.2\hsize\arraybackslash}X|}

   \hline
   & \textbf{auxiliaries} & \textbf{crops} & \textbf{pests} & \textbf{diseases} & \textbf{chemicals}  & \textbf{region} & \textbf{towns} \\	 	 	 
   \hline
   \#entries & 28  & 114 & 373  & 275 & 4968 & 26 & 33161 \\
   \hline
   \#leafs  & 28  & 103 & 334  & 241 & 4968 & 26 & 33161 \\
   \hline
   \#concepts & 0 & 18 & 53 & 40 & 0 & 0 & 0 \\
   \hline
   \#lexems  & 107 & 727 & 2673  & 1846 & 4968 & 869 & 89603 \\
   \hline
\end{tabularx} } 
\caption{Number of entries in dictionaries.}
\label{figure_statdico}
\end{figure}

\subsection[Datasets]{ Hand-crafted rule }

\begin{figure}[H]
\centering { \includegraphics[width=350pt]{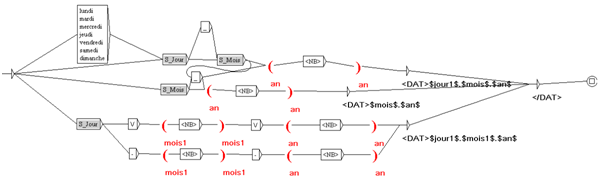} }
\caption{Local grammar (FSM) for data extraction.}
\label{figure_unitex_date}
\end{figure}

\begin{figure}[H]
\centering { \includegraphics[width=350pt]{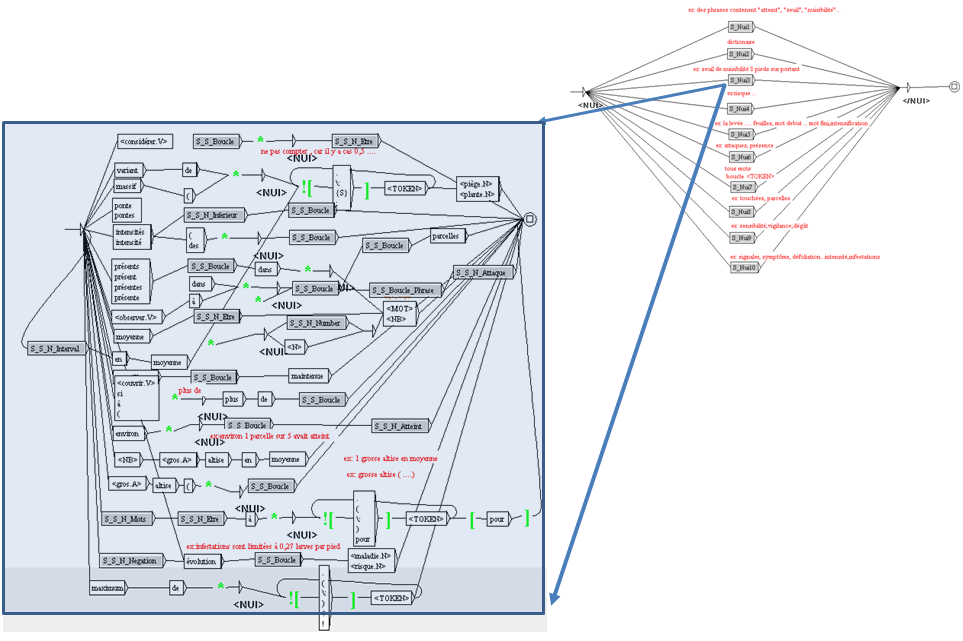} }
\caption{Local grammar (FSM) and hierarchy of FSMs for damage extraction.}
\label{figure_unitex_nuis}
\end{figure}

We used the Unitex tool (see \cite{httpUnitex}) to implement hand-crafted rules to detect some instance of named entity (date) but also contextual entities (number of a newsletter, developmental stage of a crop, intensity of damage on crop).
Graph edition emphasizes to stack and encapsulate different FSM into a more global FSM. One positive point lead to prioritise the longest matching sequence to avoid inclusion problems, hence it solves the issue to order rules execution.
Recognition of a date expression as "15 janvier 1992" or "10-2012" can be executed with the FSM showed by the following graph on Figure~\ref{figure_unitex_date}. We can see that a non-linear formulation enable unification of different schemes.
More complex and encapsulated graph enable detection of information about damage assessment (risk, prevalence or severity). 
On Figure~\ref{figure_unitex_nuis}  we see a subgraph which can detect sequence having a numerical expression about risk. This subgraph can detect an expression such as "infestations sont limit\'ees \`a 0,27 larves par pied environ 1 parcelle sur 5 avait atteint  1 grosse altise en moyenne" (infestations are limited to 0.27 larvae per foot about 1 parcel in 5 had reached  1 large flea beetle in average). It is included a global graph with 10 subgraphs.

\subsection[Datasets]{ Architecture Document Heuristics }

\newtheorem{heuristics}{Heuristics}

Organisation of a document (titles, subtitles, references, sections, headers, table, pictures, summary, introduction, discussion) can influence the way to make extraction. We call this organizatin the architecture of a document. Of course lots of architecture are availiable and the set of heuristics is not limited. We propose three heuristics can help us to be more accurate in relatioship extraction that we test with BSV and ROMEO datasets.

\begin{heuristics}  \textup{ Main entity  }  \\ 
\label{h1}
\textup{ A target entity occur in a specific title or subtitle (beginning of a paragraph or a section).  }
\end{heuristics} 

On Figure~\ref{figure_bsv} we see that main entity occur in the title of each section. It fits with heuristics~\ref{h1}.

\begin{heuristics}  \textup{ Header  }  \\ 
\label{h2}
\textup{ Different entities occurs in the header of the document (first lines).  }
\end{heuristics} 

On Figure~\ref{figure_bsv} we see that instances of entity types region, issue, date occur in the Header. It fits with heuristics~\ref{h2}.

\begin{heuristics}  \textup{ Avoid section  }  \\ 
\label{h3}
\textup{ Some paragraphs begining by a specific title can contain entity but not associated to a main entity or contextual information. }
\end{heuristics} 

\begin{figure}[H]
\centering { \includegraphics[width=350pt]{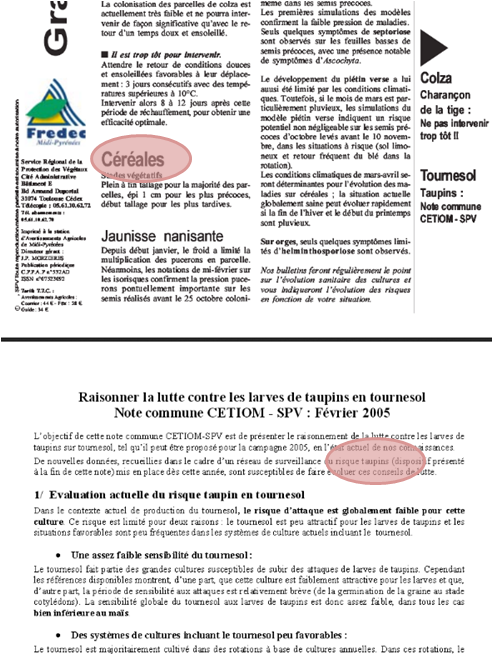} }
\caption{Neighbourhood of a section with a main entity and an avoid section for the BSV dataset.}
\label{figure_heur}
\end{figure}

On Figure~\ref{figure_heur} we see that a section begins by "Raisonner la lutte contre" ("Reasoning control against"). If we donot exclude this section of the analysis we get a relationship instance "crop/pest" as cereals/wireworms that is false. It fits with heuristics~\ref{h3}.

\subsection[Datasets]{ Unsupervised learning }

We used a classical unsupervised learning approach called cooccurrence analysis. Three family of cooccurrence can be implemented.

\begin{definition} \textup{ Entity position  }  \\ 
\label{d4}
\textup{ Let $E_i$ be a target entity. A document is split into a set of textual unit (TU). A TU can be a section, a sentence or a paragraph. Let $P_{w}^{i}$ be the position in terms of word, and $P_{TU}^{i}$ of the header word of $E_i$ in the document. We define a window by WL, i.e. the number of words at left from $P_{w}^{i}$, and WR the number of words at right from $P_{w}^{i}$. $W_R$, respectively $W_L$, can be $\infty$  if we look the right, resp. left,  context till the end, resp. the beginning, of the document. }
\end{definition} 

\newtheorem{type}{Type}

\begin{type}  \textup{ Text Unit Cooccurrence  }   
\label{t1}
\textup{ Let $E_i$ be a target entity, and $E_j$ another entity. We define the cooccurrence by the following function cooc($E_i, E_j$) is a binary function such as :}

\begin{equation}
  \label{c1}
   cooc(E_i, E_j) = \left\{ \begin{array}{ll}
         1 \  \mathrm{if \ } P_{w}^{i} \in P_{TU}^{i}\mathrm{,\ and\ }  P_{w}^{j} \in P_{TU}^{i}\mathrm{\ and\ }P_{w}^{i}\mathrm{ \ satifies\  heuristics~\ref{h1},\ ~\ref{h2}\  and\  ~\ref{h3}};   \\
        0 \ \mathrm{ else}.\end{array} \right.
\end{equation} 

\end{type}

\begin{type}  \textup{ Window Cooccurrence  }   
\label{t2}
\textup{ the same as type~\ref{t1} but now:  }

\begin{equation}
  \label{c2}
   cooc(E_i, E_j) = 1 \  \mathrm{if \ } (P_{w}^{i} - W_L)\leq P_{w}^{j} \leq (P_{w}^{i} + W_R)
\end{equation} 

\end{type}

\begin{type}  \textup{ Constrained Cooccurrence  }   
\label{t3}
\textup{ The same as type~\ref{c1} or type~\ref{c2}. But now let be a list of markers ${m_k}$,  at least one marker $m_k$ need to be located between $E_i$ and $E_j$ so : }

\begin{equation}
  \label{c3}
   cooc(E_i, E_j) = 1 \  \mathrm{if \ }  |P_{w}^{i} - P_{w}^{k}| \leq |P_{w}^{i} - P_{w}^{j}| 
\end{equation} 

\end{type}

\subsection[Step1]{ Step 1: Named Entity Recognition }\label{subsec:step1}

Assumption~\ref{a2} gives us to understand that named entities are explicitly written in the text as tokens. 
Assumption~\ref{a3} highlights important way to extract only useful named entities for a specific usage. 
In that sense extraction is dictionary-driven for better relevance (see Algorithm~\ref{algorithm1}).

\begin{table}[H]
\begin{center}
\begin{tabular}{ll}
\begin{minipage}{4.5in}
\begin{algorithm}[H]
\begin{algorithmic}[]

\medskip

\REQUIRE  dictionaries and grammars.
\ENSURE data entities.
\medskip

\STATE read\ all\ dictionaries\ (with\ nodes\ and\ leafs)\ and\ grammars

\FORALL { doc in corpus }
\FORALL { dic in dictionaries }
   \IF { words  match in dic }
   \STATE push\ data\_entities, words
   \ENDIF
\ENDFOR
\FORALL { gra in grammars }
   \IF { words match in gra }
   \STATE push\ data\_entities, words
   \ENDIF
\ENDFOR
\ENDFOR

\medskip

\STATE check\ data\ entities\ inclusive\ other\ words

\medskip

\STATE sort\_asc\ data\_entities
\FOR{ $i = 0; i<len(data\_entities)-1; i++$ }
\FOR{ $j = len(data\_entities)-1;j>i;j--$ }
   \IF { data\_entities[i] exist in data\_entities[j] }
   \IF { position of data\_entities[i] in then document is the same position of data\_entities[j] in the document }
      \STATE Remove\ data\_entities[j] 
   \ENDIF
   \ENDIF
\ENDFOR
\STATE Extract\ data\ according\ the\ entities
\ENDFOR

\medskip

\end{algorithmic}
\caption{ named entity extraction algorithm.}
\label{algorithm1}
\end{algorithm}
\end{minipage}
\end{tabular}
\end{center}
\end{table} 

There are entities for which we can not build the dictionaries, we propose the construction of grammar to extract the contents in the corpus, for instance pest significance, developmental stage of a crop, climate, location. We have integrated the Unitex tool for building grammar. Here in the following example, the uses of grammar is more reasonable, such as location, we can use the dictionary to store regions, cities or even towns. However there are names of region that combine with words of direction such as "north", "south", "west", etc. Because of that, the use of grammar will be more flexible and increase accuracy. This is some grammatical rules:

\begin{itemize}

\item \textless words in the dictionary\textgreater
\item \textless keyword 1\textgreater...\textless end of sentence or punctuation\textgreater
\item \textless words in the dictionary\textgreater...\textless keyword 1\textgreater
\item \textless keyword 1\textgreater...\textless words in the dictionary\textgreater

\end{itemize}

For example, to retrieve a developmental stage of crops in Figure~\ref{figure_grammar_stade}: "Stades: 90 \`a 100\% de couverture du sol terres color\'ees", the grammar will be as follow:

\begin{figure}[H]
\centering { \includegraphics[width=350pt]{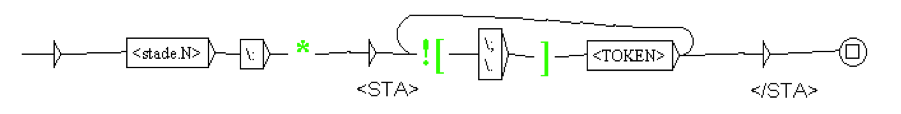} }
\caption{Local grammar (FSM) for developmental stage.}
\label{figure_grammar_stade}
\end{figure}

This diagram shows that, to begin finding a phrase that has the word "stade" or "stades" then two points, there is a loop to go through all the words in this sentence, to the meeting point signal "." or semicolon ";". Two words \textless STA\textgreater \ and \textless/STA\textgreater \ mark the result. 

\subsection[Step2]{ Step 2:  relation extraction }

Relation extraction takes as input item-sets to identify relations the export from Algorithm~\ref{algorithm1}, and plays with the three heuristics. Heuristics~\ref{h1} set that some entities are main entities (i.e. a category is a chosen as a target) and we seek relations for these entities. Heuristics~\ref{h2} sets that target entities are declared in header sections (titles, subtitles) and heuristics~\ref{h2} declares that some sections are non-relevant, we called them avoid sections and they can be specified by a beginning phrase and can end by the end of document of another phrase. Algorithm~\ref{algorithm2} describes how relation extraction is implemented.
x.ent implements also a class of algorithm to detect relation without heuristics in case, a document only consists of paragraphs (i.e a tweet, an email or a news).

\begin{table}[H]
\begin{center}
\begin{tabular}{ll}
\begin{minipage}{4.5in}
\begin{algorithm}[H]
\begin{algorithmic}[]

\medskip

\REQUIRE  data entities of the document.
\ENSURE relations.
\medskip

\FORALL {  line in this document }
             \STATE let\ paragraphs\ =\ analyze\ structure\ doc \COMMENT{this is a step in analyzing document structure  or concurrence in definition 4.}
\ENDFOR

\medskip

\FORALL { para in paragraphs }
   \IF { exists Ei and Ej in this para }
      \STATE push\ this\ relation
   \ENDIF
\ENDFOR

\medskip

\end{algorithmic}
\caption{ relation extraction algorithm.}
\label{algorithm2}
\end{algorithm}
\end{minipage}
\end{tabular}
\end{center}
\end{table}

\subsection[Step3]{ Step 3:  contextual information assignment }

Some pieces of information are considered as category of named entities because they describe a pattern of reality but they do not denote a specific object as : crop damage, developmental stage, climate. Often, these categories often cannot be designed in a dictionary but with handcrafted-rules, but they are detected at the same time as others and independently (see Section~\ref{subsec:step1}). \\

Nevertheless they can describe more precisely the context of a relationship between two entities. That is why some relationship are not only binary but n-ary (for instance crop-disease-damage). Damage here describes a magnitude of the relationship. \\

\begin{figure}[H]
\centering { \includegraphics[width=350pt]{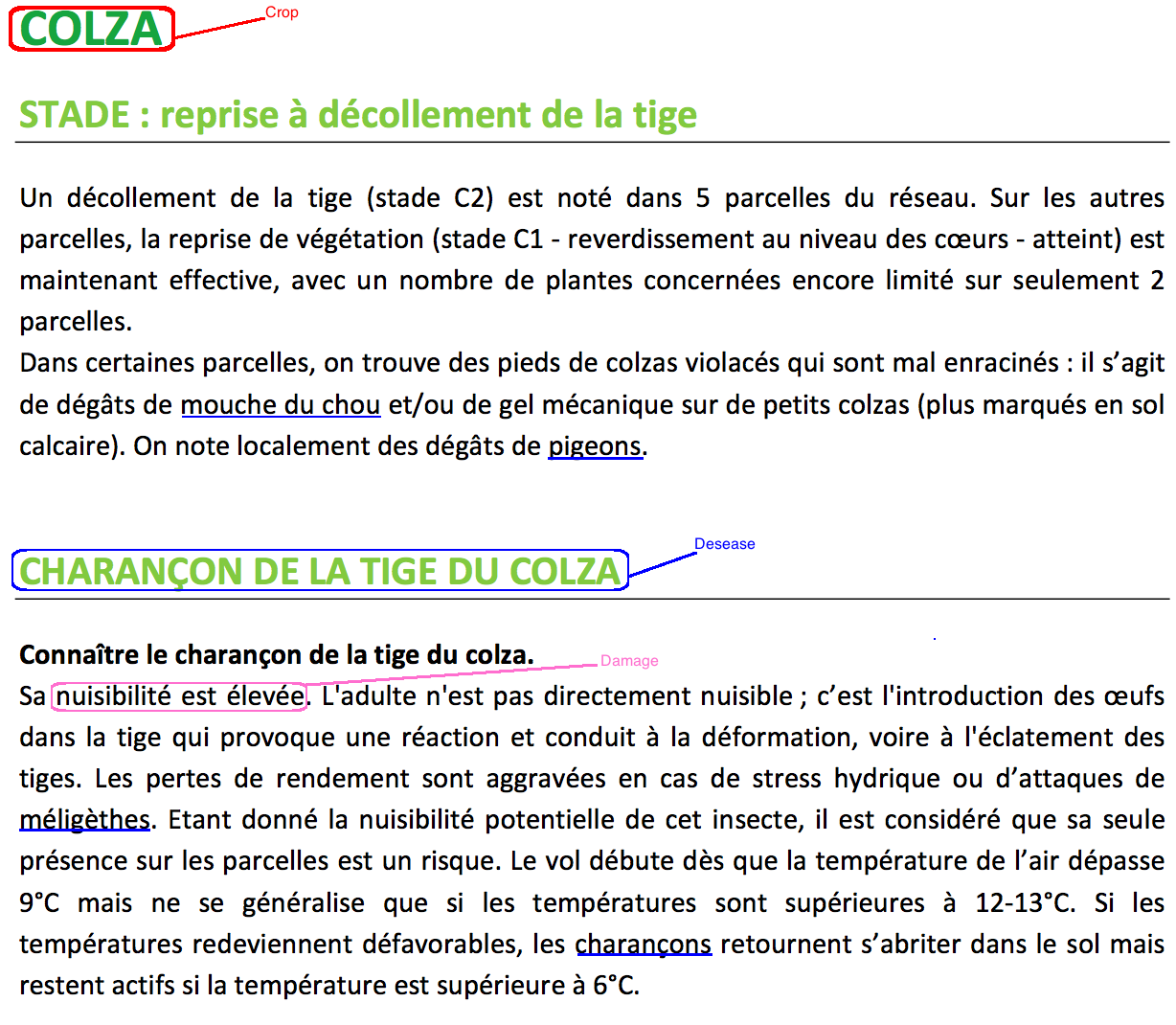} }
\caption{An example for analyzing a relationship of three entities "crop-disease-damage".}
\label{document_analysis}
\end{figure}

In the Figure~\ref{document_analysis}, we have applied an algorithm to analyze the structure of paragraphs containing the entities "crop", "disease" and "damage" to find out the relationship crop-disease-damage. The Algorithm~\ref{algorithm1} found a crop's value "Colza" at the beginning of line, this value will be the one for breaking paragraphs at this position to a position of next crop entity or to the end of document if it doesn't exists a value of crop entity at the start of line. In these paragraphs, we continue to analyze the structure of paragraphs according to disease entity, in this case "Charan\c{c}on de la tige du colza" is stated by a first line. Finally, we find all the values of damage entity in these segments that contain the values of disease entity. In this example, we will find that the relationship is "Colza:Charan\c{c}on de la tige du colza:nuisibilit\'e est \'elev\'ee" but not relation: "Colza:mouche du chou:nuisibilit\'e est \'elev\'ee".

\begin{figure}[H]
\centering { \includegraphics[width=350pt]{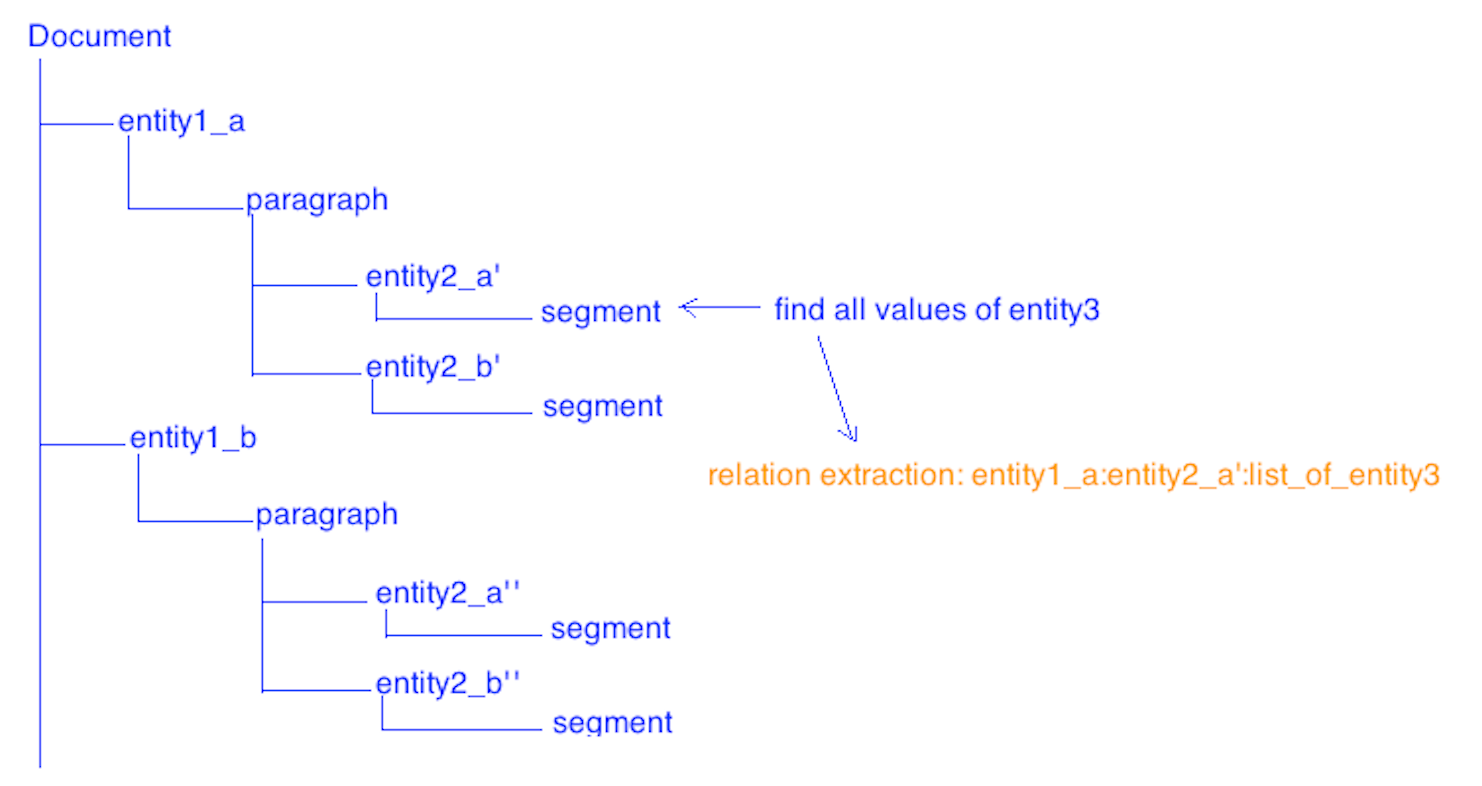} }
\caption{Analyse a document to find out a relationship of three entities.}
\label{extract_relations}
\end{figure}

\begin{table}[H]
\begin{center}
\begin{tabular}{ll}
\begin{minipage}{4.5in}
\begin{algorithm}[H]
\begin{algorithmic}[]

\medskip
  
\REQUIRE  $document, entity\_tag$
\ENSURE  paragraphs.

\FORALL { line in document}
	\STATE $mark\_curr \gets 0$
	\STATE $mark\_prev \gets 1$
	\STATE $phases \gets null$
	\IF{data of entity\_tag is the start of the line or upper case first letter in line} 
		\STATE $mark\_curr \gets mark\_curr +1$
		\IF{$mark\_curr  > mark\_prev  $}
			\STATE push\ paragraphes, phases
			\STATE $phases \gets null$
		\ENDIF
		\STATE $phases \gets phases + line$
	\ENDIF
	\IF{$length(phases) > 0$}
		\STATE push\ paragraphes, phases \COMMENT{push the final paragraph.}
	\ENDIF
\ENDFOR
\RETURN paragraphes

\medskip

\end{algorithmic}
\caption{ Transformer data.}
\label{algorithm3}
\end{algorithm}
\end{minipage}
\end{tabular}
\end{center}
\end{table} 

\begin{table}[H]
\begin{center}
\begin{tabular}{ll}
\begin{minipage}{4.5in}
\begin{algorithm}[H]
\begin{algorithmic}[]

\medskip
  
\REQUIRE  $document, entitie\_tags$
\ENSURE  relations of contextual information
\IF{$length(entity\_tags) = 3$}
	\STATE\COMMENT{check a relationship of three entities}
	\STATE $paragraphs1 = TransformerData(document,entity\_tags[0])$
	\FORALL{para1 in paragraphs1}
		\STATE $paragraphs2 = TransformerData(para1,entity\_tags[1])$
		\IF{length(paragraphs2)> 0}
			\FORALL{para2 in paragraphs2}
				\IF{$para2\ exists\ values\ of\ entity\_tags[2]$}
					\STATE $get\ values\ of\ entity\_tags[2]\ in\ para2$
					\STATE $get\ value\ of\ entity\_tags[1]\ in\ para2$ 
					\STATE $get\ value\ of\ entity\_tags[1]\ in\ para1$
					\STATE push relations, values of entities tags
				\ENDIF
			\ENDFOR
		\ENDIF
	\ENDFOR
\ENDIF
\RETURN{relations}

\medskip

\end{algorithmic}
\caption{ Contextual Information Extraction.}
\label{algorithm4}
\end{algorithm}
\end{minipage}
\end{tabular}
\end{center}
\end{table} 

\section[Results]{Results}

\subsection[Eval]{ Evaluation about extractions }\label{subsec:eval}
\begin{figure}[H]
\centering { \includegraphics[width=350pt]{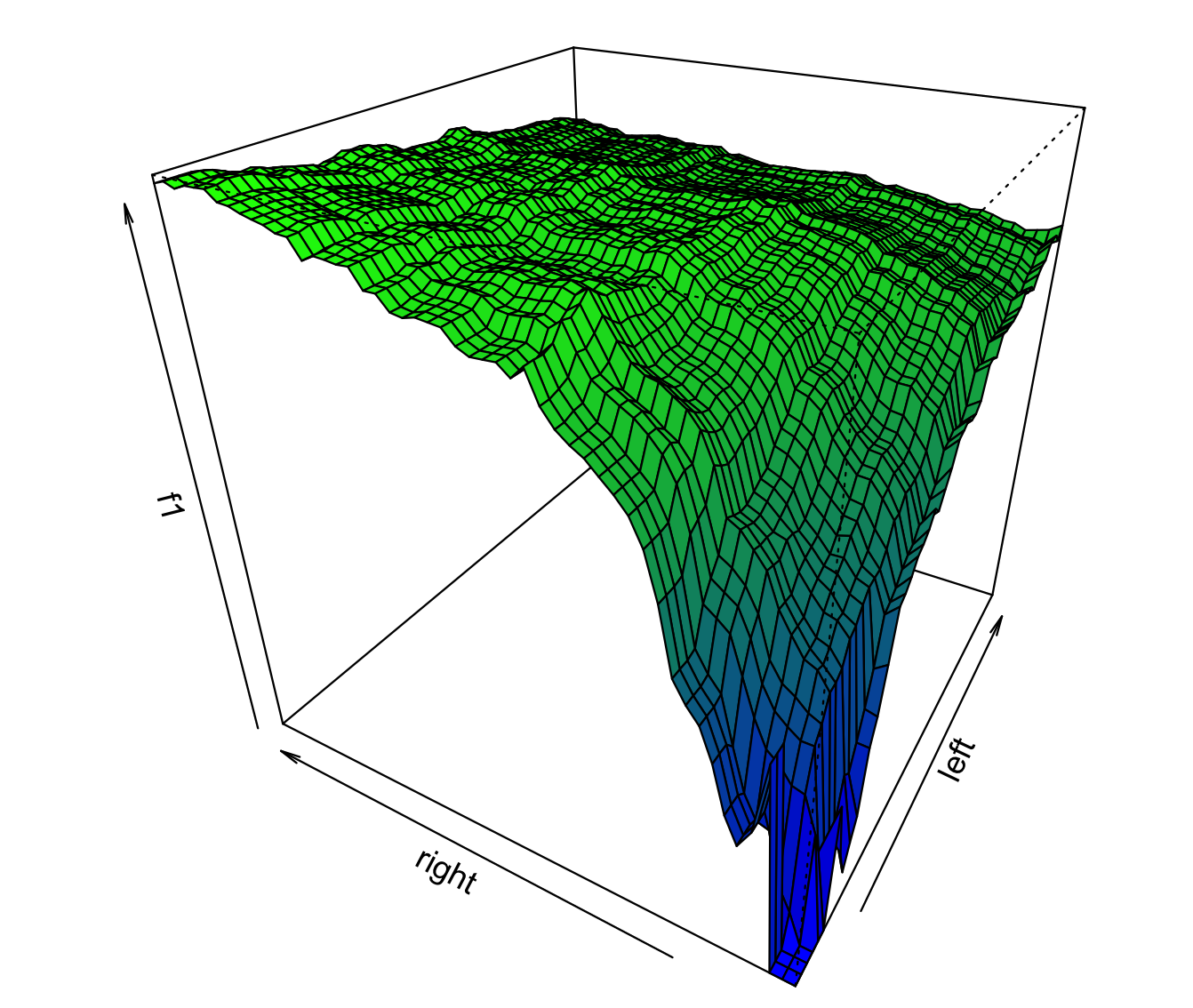} }
\caption{F1 score in the parameter space (left and right window) about cooccurrence window-based approach.}
\label{figure_f1Cooc}
\end{figure}

We proceed to a double evaluation process: \\
Firstly, we compare x.ent export of named entities with those produced by well-known approaches : exact dictionary-matching and MaxEnt approach with respectively LingPipe tool and SNER tool. Table~\ref{figure_evalNER} show results about crop, disease and pest names extraction. Standard measures to assess accuracy of a system rely on known pieces of information we aim to extract in a test dataset. The three parameters for assessment are the following: f-score (Equation~\ref{Fscore}), recall (Equation~\ref{recall}) and precision (Equation~\ref{precision}). \\
x.ent produce score  as good as those revealed by Lingpipe. Lingpipe propose also a machine learning approaches based on hidden-markov models but it gives less good results. \\
Secondly, we compared relation extraction of x.ent and those exported by SNER and cooccurrence approach with different window parameters. SNER use a parsing tree analysing and French has been considered to process BSV dataset. Table~\ref{figure_evalERR} display that x.ent capture more good relation than other state of the art approaches with a F-score about 55\%, when SNER produce 38\% and cooccurrence window-base approach 42\% (see Figure~\ref{figure_f1Cooc} about F-score variation according the window size).

\begin{eqnarray}
  \label{precision}
 0 \leq P \leq 1,     P = \frac{\#correct\_answers}{\#produced\_answers}  .    
\end{eqnarray} 

\begin{eqnarray}
  \label{recall}
 0 \leq R \leq 1,     R = \frac{\#correct\_answers}{\#possible\_correct\_answers}  .    
\end{eqnarray} 

\begin{eqnarray}
  \label{Fscore}
 0 \leq \textit{F1} \leq 1,     \textit{F1} = \frac{ \left(\beta^2 +1 \right)P.R }{ \left(\beta^2.R+P\right) } .
\end{eqnarray} 

For $Equation~\ref{Fscore}$ usually $\beta=1$.

\begin{figure}[H]
\centering

\begin{tabular}{|l|l|c|r|l|c|r|l|c|r|}
   \hline
           & \multicolumn{3}{c| }{X.ENT} & \multicolumn{3}{c| }{SNER} & \multicolumn{3}{c| }{LINGPIPE}\\
   \hline
            & P     & R     & F1    & P     & R & F1 & P & R & F1 \\
   \hline
   BIO      & 96.46 & 95.52 & 95.98 & 92.66 & 71.41 & 80.52 & 96.45 & 95.53 & 95.99 \\
   \hline
   MAL      & 96.97 & 95.53 & 96.24 & 95.46 & 77.38 & 85.38 & 96.97 & 95.52 & 96.24 \\
   \hline
   PLA      & 88.80 & 98.67 & 93.47 & 93.99 & 82.68 & 87.94 & 88.80 & 98.67 & 93.47 \\
   \hline
   REG      & 100   & 100   & 100   & 93.20 & 73.73 & 81.92 & 100 & 100 & 100 \\
   \hline
   TOT      & 94.33 & 96.67 & 95.48 & 93.68 & 76.85 & 84.41 & 94.34 & 96.65 & 95.48 \\
   \hline
\end{tabular}

\caption{Evaluation of named entity recognition.}
\label{figure_evalNER}
\end{figure}

\begin{figure}[H]
\centering

\begin{tabular}{|l|l|c|r|l|c|r|l|c|r|}
   \hline
           & \multicolumn{3}{c| }{X.ENT} & \multicolumn{3}{c| }{COOCCURRENCE}\\
   \hline
            & P     & R     & F1     & P & R & F1 \\
   \hline
   PLA-BIO  & 53.4 & 75.8 & 52.7  & 36.4 & 50.5 & 42.3 \\
   \hline
   PLA-MAL  & 58.1 & 69.5 & 63.3 & 41.3 & 38.7 & 40.0 \\
   \hline
   TOT      & 55.3 & 73.1 & 62.9 & 38.1 & 45.4 & 41.4 \\
   \hline
\end{tabular}

\caption{Evaluation of entity relationship recognition.}
\label{figure_evalERR}
\end{figure}

\subsection[Visualisation]{ Visualisation }

The x.ent tool has been developped with \emph{Perl} modules concerning the parsing function and but is encapsulated under an \emph{R} package availaible on the \emph{R} platform (see \cite{R-project:2015}). The package offers also \emph{R} functions to explore results of extraction : parallel coordinates, histogram, Venn diagram, stacked bar graph and statistical test on pairwise relation.

\begin{figure}[H]
\centering { \includegraphics[width=350pt]{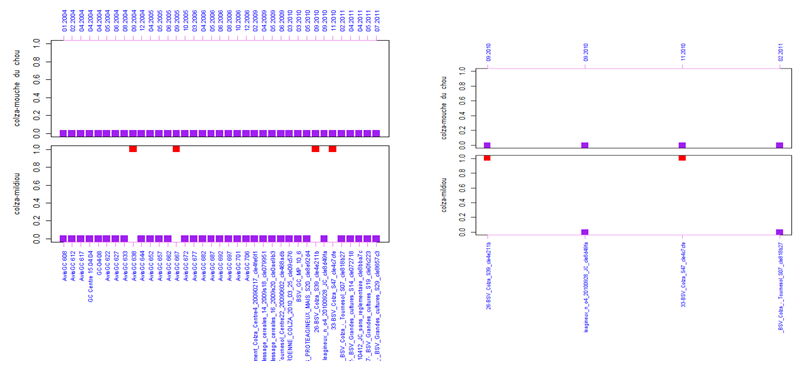} }
\caption{Parallel coordinate display.}
\label{figure_display_parallel}
\end{figure}

On Figure~\ref{figure_display_parallel} we see an example of parallel coordinate visualization between two sets of entities (e1 and e2). e1 is a target entity with which we seek relations. About BSV dataset the target entityt is crop category. In the example e2 are a set of entity from different categories ("mouche du chou" is an instance of pest category and "mildiou" is an instance of disease category).  \\
The R code is the following: \\
\hspace *{1cm} \textit{ xplot(e1="colza",e2=c("mouche du chou", "mildiou")) } \\
We can add a constraint about the time : \\
\hspace *{1cm} \textit{ xplot(e1="colza",e2=c("mouche du chou", "mildiou"),t=c("09.2010","02.2011")) }

\begin{figure}[H]
\centering { \includegraphics[width=200pt]{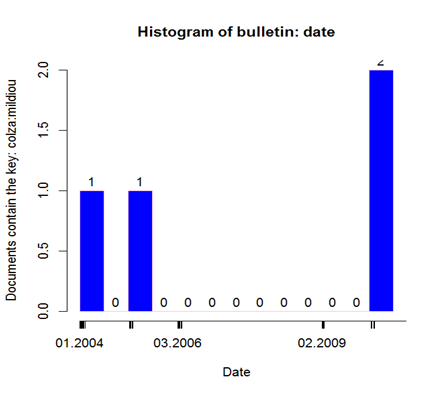} }
\caption{Histogram over time of a relation.}
\label{display_histo}
\end{figure}

Figure~\ref{display_histo} shows the distribution over time about a specific relation "colza:mildiou", but it could work with any instance of an entity but only in the case a date entity is extracted from the dataset. \\
The R code is the following: \\
\hspace *{1cm} \textit{ xhist("colza:mildiou") } 

\begin{figure}[H]
\centering { \includegraphics[width=350pt]{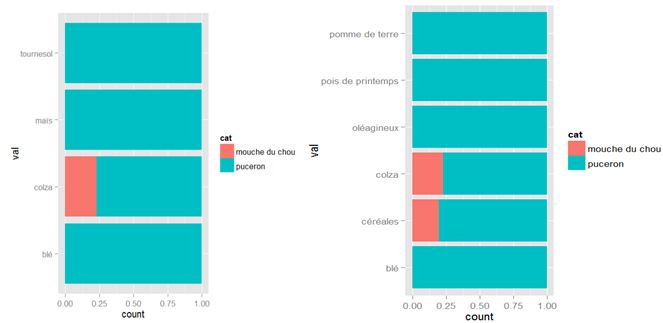} }
\caption{stacked bar graph.}
\label{figure_display_stacked}
\end{figure}

Figure~\ref{figure_display_stacked} shows  and a stacked bar graph representing for a first set of entities, in the example the crops: "bl\'e", "ma\"is", "tournesol", "colza", the proportion of each instance of entity of the second set, hereafter "mouche du chou", "puceron". \\
The R code is the following: \\
\hspace *{1cm} \textit{ xprop( c("bl\'e", "ma\"is", "tournesol", "colza") , c("mouche du chou", "puceron") ) } \\
If the first set the the whole set of instance of the target entity catagory (hereafter instances of crops) and having at least 2 occurences : \\
\hspace *{1cm}\textit{v1 = as.vector(xdata\_value("p")\$value[xdata\_value("p")\$freq \textgreater 2])}\\
\hspace *{1cm}\textit{xprop(v1,c("mouche du chou","puceron")) } 

\begin{figure}[H]
\centering { \includegraphics[width=200pt]{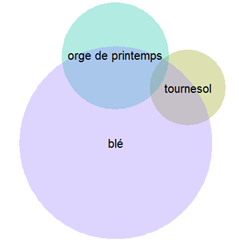} }
\caption{Venn diagram.}
\label{figure_display_venn}
\end{figure}

Figure~\ref{figure_display_venn} shows a Venn diagramme between a set of instance from the target category (hereafter "bl\'e","orge de printemps","tournesol") and a set of instances from specified categories (hereafter b and m, denotating respectively pest and disease): \\
The R code is the following: \\
\hspace *{1cm} \textit{ xvenn(v=c("bl\'e","orge de printemps","tournesol"),e=c("b","m")) } 

Figure~\ref{figure_tests} shows a comparison between a crop (hereafter "bl\'e") and all possible instance of another entity category (hereafter instances of pest category). Four tests has been implemented for the function : Kolmogorov, Wilcoxon, Student and GrowthCurves. At moment no decision function makes interpolation over the tests to decide if yes or no the p-values agree for a positive similarity or not. Figure~\ref{figure_saturation} shows an export with all p-values saturating at 1. For instance "bl\'e:limace des jardins" and "bl\'e:adventice" have the same distribution across the BSV dataset. It means that "limace des jardins" occurs at same time that "adventice" in "bl\'e" crop cultures. \\
The R code is the following: \\
\hspace *{1cm} \textit{ xtest( "bl\'e", as.vector(xdata\_value("p"))) } 

\begin{figure}[H]
\centering
\renewcommand{\arraystretch}{1.2}
{\scriptsize

\begin{tabularx}{12.5cm}{|>{\centering\hsize=0.2\hsize\arraybackslash}X|
>{\centering\hsize=1.8\hsize\arraybackslash}X|
>{\centering\arraybackslash}X|
>{\centering\hsize=1.2\hsize\arraybackslash}X|
>{\centering\hsize=0.8\hsize\arraybackslash}X|
>{\centering\arraybackslash}X|}

   \hline
            & relation                                          & KOLMOGOROV      & WILCOXON     & STUDENT      & GrowthCurves  \\
   \hline
   700      & bl\'e:m\'elig\`ethe/bl\'e:thrips                          & 1.00            & 0.13         & 0.13         & 0.02  \\
   \hline
   543      & bl\'e:cicadelle/bl\'e:pyrale                          & 1.00            & 0.00         & 0.00         & 0.02  \\
   \hline
   613      & bl\'e:crioc\`ere/bl\'e:thrips                           & 1.00            & 0.00         & 0.00         & 0.02  \\
   \hline
   689      & bl\'e:m\'elig\`ethe/bl\'e:puceron des \'epis de c\'er\'eales    & 0.91            & 0.00         & 0.00         & 0.02  \\
   \hline
\end{tabularx}  } 

\caption{Pairwise relations comparison.}
\label{figure_tests}
\end{figure}

\begin{figure}[H]
\centering
\renewcommand{\arraystretch}{1.2}
{\scriptsize

\begin{tabularx}{12.5cm}{p{12.1cm}}

   \hline
bl\'e:adventice/bl\'e:limace des jardins \\
   \hline
bl\'e:adventice/bl\'e:puceron des c\'er\'eales et du rosier \\
   \hline
bl\'e:campagnol des champs/bl\'e:corbeau freux \\
   \hline
bl\'e:campagnol des champs/bl\'e:pyrale \\
   \hline
bl\'e:campagnol des champs/bl\'e:zabre des c\'er\'eales \\
   \hline
bl\'e:c\'ecidomyie jaune du bl\'e/bl\'e:charan\c{c}on \\
   \hline
bl\'e:c\'ecidomyie jaune du bl\'e/bl\'e:charan\c{c}on de la tige \\
   \hline
bl\'e:c\'ecidomyie jaune du bl\'e/bl\'e:mouche grise des c\'er\'eales \\
   \hline
bl\'e:c\'ecidomyie jaune du bl\'e/bl\'e:noctuelle \\
   \hline
bl\'e:c\'ecidomyie jaune du bl\'e/bl\'e:oscinie de l'avoine \\
   \hline

\end{tabularx}   } 
\caption{Global saturation of all tests.}
\label{figure_saturation}
\end{figure}

\subsection[Integration]{ Integration }

The x.ent tool has been used to parse BSV dataset so as to export result in a csv format and used in a database management system to be queried by end-users. In this system relations are pivotal information to offer useful piece of information for information retrieval. We can mention four cases of usage with main query and refinement:
\begin{itemize}

\item main query is relation crop-disease, refinement damage and region on map. \\
example: Crop=wheat, Disease=rust, On map : risk assessment, region=Burgondy

\item   main query is crop, refinement pest (relation crop-pest) and region on map. \\
example: Crop=rapeseed, On map : Pest=cabbage maggot, region=Centre, and document sorting by date

\item   main query is disease, refinement pest (relation crop-disease) and region on map. \\
example: disease=potato late blight, On map : crop=potato, region=Burgundy

\item   main query is pest, refinement pest (relation crop-pest) and region on map. \\
example: pest=fly, On map : crop=wheat, region=Midi-Pyr\'en\'ees

\end{itemize}

At present 4 users specialists about potato and wheat take benefit from the database and platform-as-service. More epidemiologists and agronoms, from the integrated crop protection network (r\'eseau PIC - protection int\'egr\'ee des cultures) including 400 subscribers, are potentially interested in using this web platform. Risk analysis in a sociological point of view is also possible.

\begin{figure}[H]
\centering { \includegraphics[width=350pt]{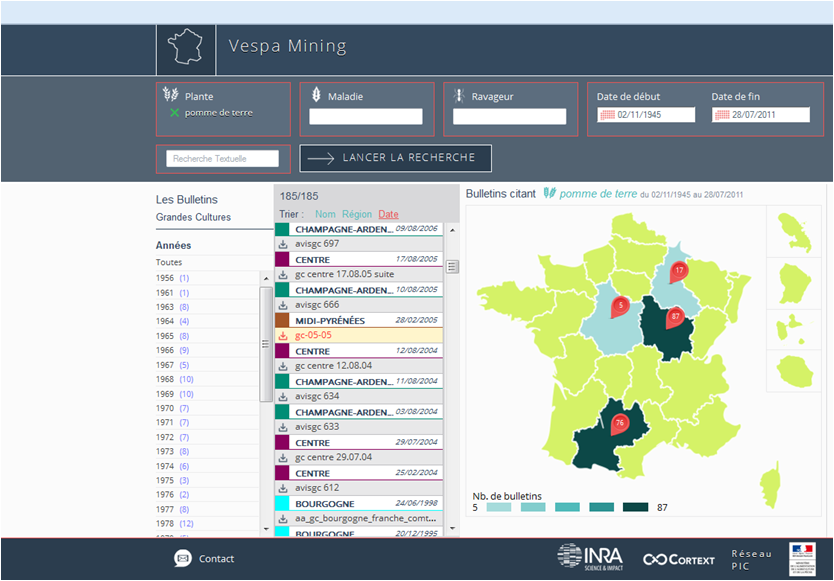} }
\caption{Vespa user interface.}
\label{figure_VespaMining}
\end{figure}

\subsection[Program availability]{ Availability }

We developed, improved and applied a relation extraction method we implemented as an \emph{R}-project package (\emph{x.ent}). The package is available from the CRAN \emph{R} project server (http://cran.r-project.org/ see Software, Packages; \emph{x.ent} version 1.0.6), and downloadable from the \emph{R} graphical user interface (required \emph{R} libraries : \emph{xtable}(see \cite{Dahl:2014}), \emph{jsonlite} (see \cite{Ooms:2014}), \emph{venneuler} (see \cite{Wilkinson:2014}), \emph{ggplot2} (see \cite{Wickham:2014}), \emph{stringr} (see \cite{Wickham2:2014}), \emph{opencpu}(see \cite{Ooms2:2014}) and \emph{rJava} \cite{rJava:2014} ). \\
The results of BSV corpus processing has been stored in a relational database with a web-front web access. The temporary website http://vespa.cortext.net display the front-end interface in which a user can query the result to retrieve relevant documents.

\subsection[Discussion]{ Discussion }

Information extraction is not a new field but new opportunity with new usages and new corpora emphasizes this kind of task. 
If number of document in a corpus can not be huge (several ten thousands to several millions), the number of possible relations has no limit. Extract good and relevant relations, store all relations, and query relations in a concrete usage context can be challenges. \\
Lots of factors can influence relation extraction. We explore the capacity of syntactic expression in document to extract relations. Indirect relation are also possible, as in genetics when a genecist set that geneA interact geneB and geneB interact geneC than geneA can be in interaction with geneC, of if geneA interact with geneC in a species, then it is also a putative relation in another species. In our BSV dataset we do use any inference protocol. We try to take into account all possible signs in a document. Hence our approach goes further than a linguistic approach the aim of which is to analyze the structure of each sentence. Our point of view is equivalent to argumentative analysis when part of speech are linked by sections. In this point of view we show that sometimes specific concept of interest can be situated in a special location in a document as in a text of theater about speaking characters, or in a newsletter with titles. \\

\section[Conclusion]{Conclusion}

We developed, improved and applied a relation extraction method available as a R package. The tool has been involved into an information system called Vespa Mining with end-user (agronoms and epidemiologists). \\
Extraction task involve the user to design a proto-ontology of its domain with a set of categories. Each category make sens with instances (string sequences) for which small local grammars and flat dictionaries can fit in documents. A target category is settle to search other instances of another category as a relationship and contextual information.
The tool relies on both hypothesis that named entities are extractible and that document structure helps extraction. We compare with state of the art tool and we show that if x.ent can reach the performance of named entity recognizers, assessment about relation extraction give better scores. Exploitation of document structure together with unsupervised learning can achieves high score of extraction. \\
We used two datasets. A literary dataset about a Shakespeare theater piece and an agricultural newsletter dataset. The goal about the newsletter was to learn relations as crop-disease-damage and crop-pest-damage. Designing an evaluation dataset  we obtain an F-score $\approx$ 55. \\
Our interest was also to help a user to explore the large potential amount of relationships. Two means was implemented in that direction. Firstly information visualization capacities such as : parallel coordinates, histogram, Venn diagram, stacked bar graph and statistical test on pairwise relation. Secondly an integration of the tool in a user-friendly platform with Concrete real-world information. Here the user can browse the dataset through relationships and complementary information (locations, damage magnitudes, or simple keywords) through geolocatlisation and feedback to original documents.

\section*{Acknowledgments}
Special thanks  to Kurt Hornik (Vienna University) for its discussion about technical aspects; and to Roselyne Corbi\`ere (INRA - Rennes center) and Vincent Cellier (INRA - Dijon center) about comments on usage with the web platform; and to Jean-Noel Aubertot (INRA - Toulouse center) for its initiative about the BSV dataset construction. The methodology discussed in this paper has been supported by the VESPA grant from the French Ministry of Agriculture and BSN5 grant from French Ministry of National Education and Research.

\bibliographystyle{plain}
\bibliography{submission_turenne_phan}

\end{document}